\documentclass[letterpaper, 10 pt, conference]{ieeeconf}  

\IEEEoverridecommandlockouts                              

\usepackage{cite}
\usepackage{amsfonts}
\usepackage{graphicx}
\usepackage{multirow}
\usepackage[normalem]{ulem}
\usepackage{booktabs}
\usepackage{url}

\usepackage{color}
\usepackage{xcolor}

\usepackage{CJKutf8}

\title{\LARGE \bf
Jointly Optimized Global-Local Visual Localization of UAVs
}

\author{Haoling Li$^{1}$, Jiuniu Wang$^{2}$, Zhiwei Wei$^{3}$ and Wenjia Xu$^{4 *}$
\thanks{Work done during Haoling Li is an undergraduate at Beijing University of Posts and Telecommunications supervised by Wenjia Xu.}
\thanks{ * Corresponding author: Wenjia Xu}
\thanks{$^{1}$Haoling Li is with Tsinghua Shenzhen International Graduate School, Tsinghua University, Shenzhen 518055, China.}%
\thanks{$^{2}$Jiuniu Wang is with China City University of Hong Kong, 999077, Hong Kong, China.}
\thanks{$^{3}$Zhiwei Wei is with Aerospace Information Research Institute, CAS, Beijing 100094, China.}
\thanks{$^{4 *}$Wenjia xu is with the State Key Laboratory of Networking and Switching Technology, Beijing University of Posts and Telecommunications, Beijing 100876, China
{\tt\small xuwenjia@bupt.edu.cn}}
}

\begin{document}

\maketitle
\thispagestyle{empty}
\pagestyle{empty}

\begin{abstract}

Navigation and  localization of UAVs present a challenge when global navigation satellite systems (GNSS) are disrupted and unreliable. Traditional techniques, such as simultaneous localization and mapping (SLAM) and visual odometry (VO), exhibit certain limitations in furnishing absolute coordinates and mitigating error accumulation. 
Existing visual localization methods achieve autonomous visual localization without error accumulation by matching with ortho satellite images. However, doing so cannot guarantee real-time performance due to the complex matching process. To address these challenges, we propose a novel Global-Local Visual Localization (GLVL) network. Our GLVL network is a two-stage visual localization approach, combining a large-scale retrieval module that finds similar regions with the UAV flight scene, and a fine-grained matching module that localizes the precise UAV coordinate, enabling real-time and precise localization. The training process is jointly optimized in an end-to-end manner to further enhance the model capability. Experiments on six UAV flight scenes encompassing both texture-rich and texture-sparse regions demonstrate the ability of our model to achieve the real-time precise localization requirements of UAVs. Particularly, our method achieves a localization error of only 2.39 meters in 0.48 seconds in a village scene with sparse texture features.

\end{abstract}

\section{INTRODUCTION}

The acquisition of precise location coordinates is essential for Unmanned Aerial Vehicles (UAVs), and most of the UAVs utilize the Global Navigation Satellite System (GNSS) to provide coordinates \cite{gyagenda2022review}. However, the accurate position of the UAV cannot be obtained in GNSS-denied or GNSS-spoofed environments caused by multi-path interference and electromagnetic interference, etc \cite{balamurugan2016survey}.  Consequently, autonomously localizing the UAVs without the help of electromagnetic signals assumes paramount significance \cite{kaleem2018amateur}. Recent years have witnessed the emergence of diverse autonomous localization techniques tailored for unmanned systems, encompassing visual localization, geomagnetic localization, and often costly inertial navigation systems \cite{couturier2021review}.

Within the realm of UAV autonomous navigation, a visual localization system that achieves comparable accuracy would be suitable as it is affordable and can replace GNSS when there are signal issues. Simultaneous Localization and Mapping (SLAM) \cite{dissanayake2001solution} and Visual Odometry (VO) \cite{nister2004visual} are capable of ascertaining the relative spatial disposition and the motion properties. SLAM technology identifies previously visited locations by constructing maps of the surrounding environment, which requires a closed-loop detection solution, making it insufficient to meet the stringent localization requirements of UAVs navigating in unknown and expansive outdoor terrains \cite{mur2015orb}. VO can overcome the problems of SLAM by utilizing optical flow techniques to estimate the relative displacement and orientation of a UAV \cite{nister2004visual}, while cannot provide absolute latitude and longitude coordinates for the UAVs when the initial location is unknown. 
In addition, the attendant localization errors of VO systems tend to accumulate with the distance travelled progressively \cite{bloesch2015robust}. 

Given the aforementioned constraints, it becomes crucial to develop a visual localization system capable of delivering precise latitude and longitude coordinates for UAVs in outdoor environments \cite{warren2018there}. With the development of remote sensing technologies, this idea is further motivated by the fact that we now have access to high-resolution satellite images covering virtually any location on Earth, each pixel meticulously annotated with exact coordinates \cite{leprince2007automatic}. Consequently, aligning the ground view captured by UAV images with the corresponding ortho satellite images enables the immediate extraction of coordinates \cite{mantelli2019novel}. Previous works involved patching the large-scale ortho satellite images into smaller local regions and retrieving the UAV ground images, offering a rough coordinate estimate \cite{nguyen2014integrating}. Conversely, alternative approaches conduct detailed point-to-point matching between the large-scale ortho satellite images and UAV ground images. While this yields relatively precise localization outcomes, it is notably time-intensive and falls short of meeting real-time demands \cite{shan2015google}.

\begin{figure}[!t]
\centering
	\includegraphics[width=\linewidth]{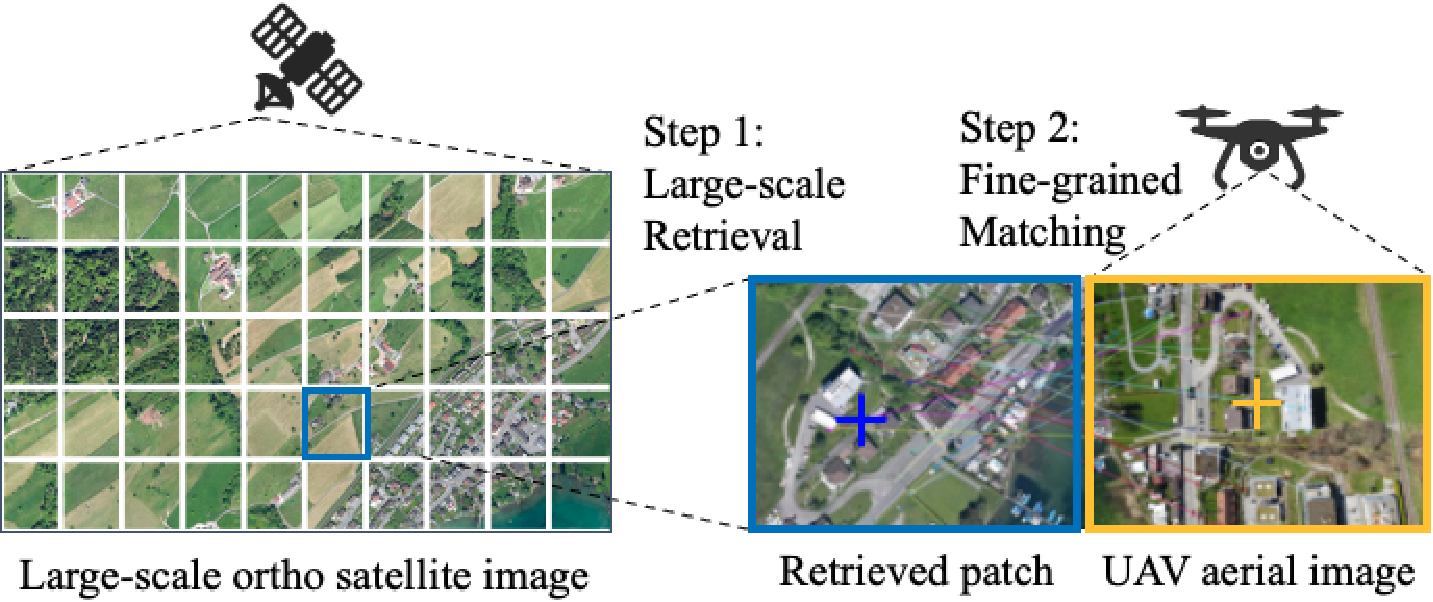}
	\caption{Our two-stage UAV visual localization strategy which is both time-efficient and accurate.}
    \label{fig:teaser_fig}
\end{figure}

To migrate these problems, we propose a novel Global-Local Visual Localization~(GLVL) Network to perform time-efficient and precise two-stage visual localization, as shown in Figure~\ref{fig:teaser_fig}. In the first stage, given a UAV frame, we utilize our Large-scale Retrieval module to retrieve visually similar image patches from the large-scale ortho remote sensing images and quickly estimate the approximate location of the UAV within a vast geographic area. In the second stage, we execute point-to-point matching between the retrieved similar image patches and the UAV image through our Fine-grained Matching module, then obtain the accurate coordinate of the UAV with homography transformation.

Retrieving correct remote sensing image patches that cover the same location with UAV images requires concentration on both the global similarity and the distinct local landmarks, especially in areas with low texture, like villages. 
Besides, the pivotal challenge of extracting distinctive regional image features and local keypoint features assumes paramount significance to the precision of the matching processes.
To this end, we have developed an end-to-end model that integrates both large-scale retrieval and fine-grained matching stages, which enables us to fully extract the global and local visual information. This combination enables the representation learning framework to emphasize both the regional feature similarities and the salient information in local keypoints.

To sum up, our work makes the following contributions:  
\begin{enumerate}[]
\item We introduce a two-stage UAV visual localization strategy, designed to expeditedly determine the area where the UAV is located and pinpoint its localization.
\item We develop an end-to-end representation learning network with the ability to concurrently attend to the similarity of local structural features and concern keypoint information, thereby augmenting retrieval and matching accuracy via joint training strategies.
\item Extensive experiments are conducted on six scenes, encompassing both texture-rich and texture-sparse regions, to validate the effectiveness and generalization ability of our proposed method. Specifically, our method achieves a localization error of just 2.39 meters in a mere 0.48 seconds, whereas other state-of-the-art methods yield a localization error of 7.06 meters.
\end{enumerate}

\section{RELATED WORKS}

\subsection{SLAM and Visual Odometry}

SLAM-based UAVs have been used in challenging indoor environments, such as sewers, mines and nuclear power plants, where human access is limited \cite{tiemann2018enhanced, moura2021graph}. With advancements in computing devices and sensing technologies, SLAM technology has been applied in an expanding range of outdoor settings \cite{milford2011aerial, ait2017outdoor, shao2021monocular}.
VO estimates the motion and position of an object using visual data from cameras or other sensors. It involves tracking the movement of features in the environment over time to infer object motion \cite{guizilini2011visual}. 
 
SLAM and VO can be combined to enhance localization and map building, providing a more comprehensive and precise representation of the environment \cite{sonugur2022review}. 
Integrating multiple sources of information can achieve a more accurate and robust localization. However, both SLAM and VO techniques are unable to detect the coordinates of UAVs, which can result in a growing localization error over time throughout the flight range. This limitation can be addressed by incorporating additional sensors to provide absolute positioning information.

\subsection{Remote Sensing Image Matching}
Remote sensing image matching techniques are employed to achieve accurate visual geo-localization of UAVs by comparing images captured by the UAV camera with a reference remote sensing image that contains precise geolocation coordinates. These methods are unaffected by cumulative errors and offer precise positioning information for long-endurance UAVs operating in outdoor environments \cite{couturier2021review}. 

Traditional methods rely on template matching and feature point matching to achieve image matching across different sources. However, with the widespread adoption of Convolutional Neural Networks (CNNs) in various computer vision tasks, current approaches increasingly utilize deep learning methods to address visual localization problems \cite{amer2017convolutional, nassar2018deep, goforth2019gps, yan2022crossloc, sarlin2023orienternet}. 
To address the significant difference in geographical coverage between the reference remote sensing image and the image acquired by the UAV camera, certain techniques employ image retrieval to narrow down the search area \cite{berton2022deep}.

Our approach draws inspiration from the exceptional performance of SuperPoint \cite{detone2018superpoint} in various vision tasks and concentrates on precisely matching local feature points. In contrast to previous methods, our technique enables autonomous geo-localization.

\section{METHODS}

\begin{figure*}[htbp]
\centering
	\includegraphics[width=\linewidth]{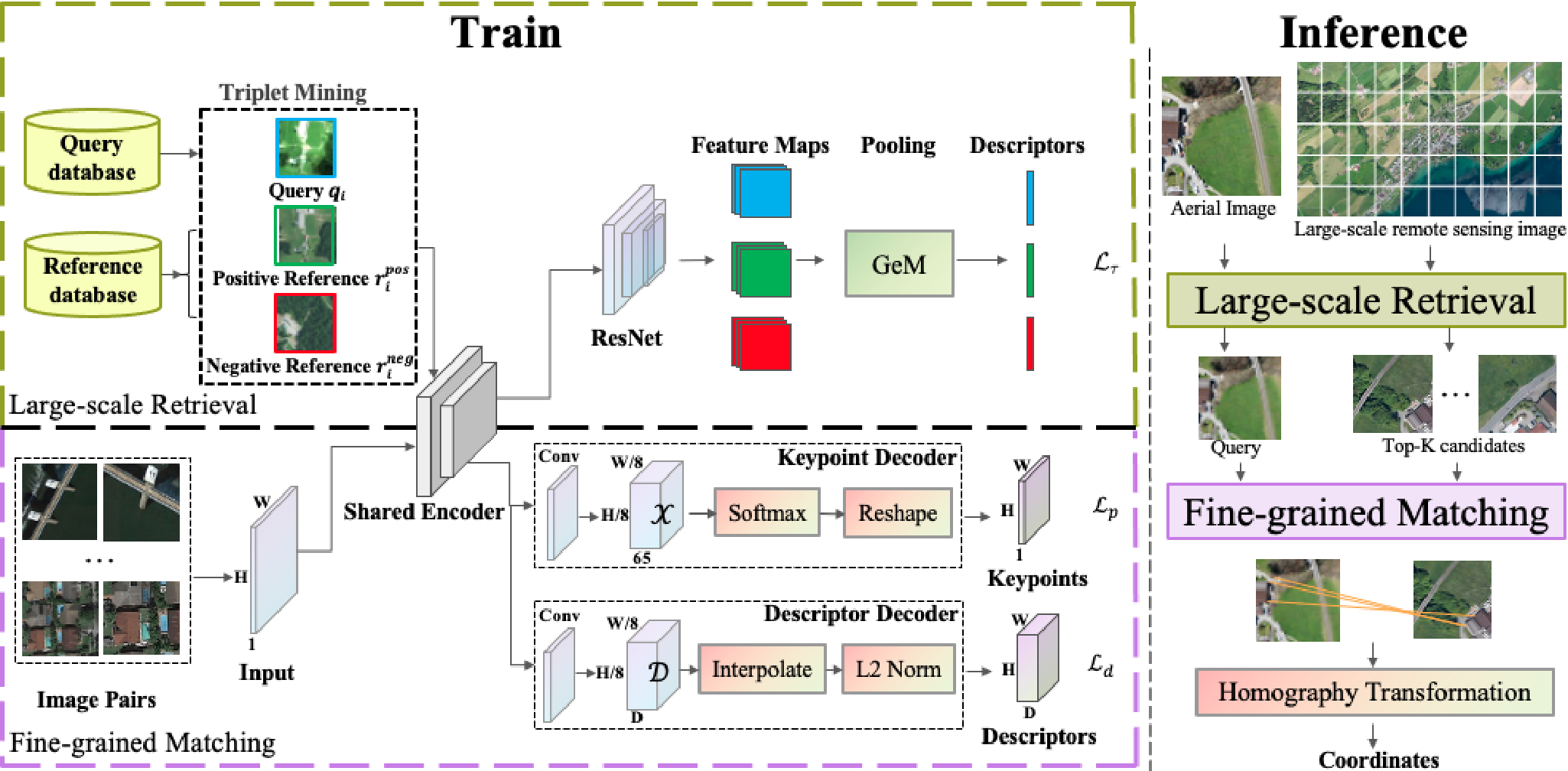}
	\caption{The train and inference process of our Global-Local Visual Localization network which combines Large-scale Retrieval and Fine-grained Matching.}
    \label{figframework}
\end{figure*}

Large-scale UAV visual localization must effectively balance the requirements of accuracy and efficiency within UAV navigation systems. 
To address this challenge, GLVL reduces the search space by image retrieval, and then GLVL performs keypoint detection and matching  between aerial image and satellite image patch. 
The precise localization of the UAV is achieved through homography transformation. Notably, GLVL offers end-to-end pixel-level matching of image pairs, eliminating the need for complex post-processing. In this section, we provide a comprehensive and detailed introduction to the training and inference processes of GLVL's Large-scale Retrieval and Fine-grained Matching modules, which are shown in Figure \ref{figframework}.

\textbf{Problem Definition.} 
The UAV visual localization aims to obtain the position of the UAV through image matching between the satellite image and the aerial image. Given $n$ aerial images $\mathbb{Q}=\{q_1, q_2, \dots , q_n\}$, the task could be formulated as finding their positions $\mathbb{P} = \{p_1, p_2, \dots , p_n\}$ given a large-scale ortho satellite image $\mathbb{M}$ where each pixel is annotated with longitude and latitude coordinates. These $n$ aerial images are usually taken continuously during the flight of the UAV, and the ground-truth UAV coordinate of each frame $q_i$ is denoted as $p_i^{coords}$, indicating the UAV's current position.
These satellite images involve GPS coordinates, $LL_{coords}$ and $UR_{coords}$, representing the lower left and upper right corners of the map, respectively. Additionally, the map size is denoted as $M_w \times M_h$. Finally, we calculate the position gap between the ground-truth UAV coordinates, $p_i^{coords}$, and the predicted coordinates, $p_{i}$.

\subsection{Train}

\textbf{Large-scale Retrieval.} 
We adapt a Large-scale Retrieval module to get the approximate range of the $i$-th aerial image $q_i$. The satellite image $\mathbb{M}$ is cropped into $m$ patches as:
\begin{equation}
\mathbb{R} = \{ r_1,  r_2, \dots , r_m \}    \,,
\end{equation}
where all pixels in $\mathbb{M}$ are included in $\mathbb{R}$, and there could be some overlap between $r_j$ and  $r_k$ (here $j \neq k$).

The retrieval $\mathrm{Ret}(\cdot)$ aims to obtain the  candidate patches which are highly related to the aerial image $q_i$ as:
\begin{equation}
\mathbb{R}_i = \mathrm{Ret}(\mathbb{R}, q_i)   \,,
\end{equation}
where $\mathbb{R}_i \subset \mathbb{R}$.

In the image retrieval task, visual features such as shape, colour, texture, and spatial information are compared to measure the similarity between the query image and the reference database image. However, when dealing with satellite images and aerial images, the retrieval task faces a significant challenge due to the feature variability of heterogeneous images. This variability is further amplified by the time difference between satellite image production and UAV aerial image acquisition. 
Based on the aformentioned consideration and analysis, we have leveraged convolutional neural networks in our research due to their exceptional recognition capabilities, compact representation, and high search efficiency.

 The detailed structure of the module is shown in Figure \ref{figframework},  where we initially extract the triplet $\mathcal{T}=\left \langle q_i, r^{pos}, r^{neg} \right \rangle$ from the training data. Here, $r^{pos}$ and $r^{neg}$ represent the positive and negative reference images of $q_i$, respectively.
To optimize computational resources, we implement a shared encoder for two modules, which not only conserves computational power but also enhances the encoder's performance through iterative training of the two modules. Subsequently, we have employed ResNet \cite{he2016deep} as the backbone to extract feature maps from image triplet.
To address the memory consumption issue arising from high output dimensionality, we incorporate a lightweight pooling layer GeM \cite{radenovic2018fine}, which facilitates the acquisition of a more compact feature descriptor.
We define the triplet loss function for each sample in the retrieval module as:
\begin{equation}
\mathcal{L_T} (q_i, r_p,  r_n) = \max \{d(q_i, r_p) - d(q_i, r_n ) + \delta, 0\}    
\end{equation}
where $d$ refers to the Euclidean distance between the two feature descriptors, $\delta$ is the margin hyperparameter. So that we got the $K$ candidate patches $\mathbb{R}_i = \{ r^{pos}_1,  r^{pos}_2, \dots , r^{pos}_K \}$.

\textbf{Fine-grained Matching}. 
After reducing the search space of large-scale satellite images, the Fine-grained Matching module proficiently detects and accurately describes the keypoints of the retrieved candidate satellite image patches. The module $\mathrm{Fgm}(\cdot)$ would match the points between the aerial image $q_i$ and the candidate patch $r^{pos}_k$ as:
\begin{equation}
S_{i,k} = \mathrm{Fgm}(q_i, r^{pos}_k) \,,   
\end{equation}
 where $S_{i,k} = \{ (C_1, C'_1),\dots , (C_L, C'_L)\}$ is the set of point pairs, $C_l$ is the point from $q_i$  and $C'_l$ is the point from $r^{pos}_k$ ($l \in [1, L]$). 
SuperPoint leverages a self-supervised training strategy to achieve pixel-level keypoint detection and feature description, surpassing the capabilities of traditional angle detectors such as SIFT \cite{lowe2004distinctive} and ORB \cite{rublee2011orb}.

In the Fine-grained Matching module of Figure \ref{figframework}, the shared encoder serves as the encoder component of the module. 
It is then followed by the Keypoint Decoder and Descriptor Decoder, which align with the vanilla SuperPoint architecture.
For Keypoint Decoder, the dimension of $\mathcal{X}$ is $\frac{H}{8} \times \frac{W}{8} \times 65$, representing the response value of the keypoint on the corresponding pixel, which is the probability value after Softmax calculation. The dimension of the output tensor is $H\times W$, which is the size of the image. Here 65 denotes a localized region of $8\times8$ pixels of the original image, plus a non-featured point dustbin.
The dimension of the output tensor is changed by doing Reshape of the image. The loss function of Keypoint Decoder is:
\begin{equation}
    \mathcal{L}_{p}(\mathcal{X}, Y)=\frac{1}{H \times W} \sum_{} l_{p}\left(\mathbf{x} ; y\right),
\end{equation}
where Y is the labeled keypoint corresponding to $\mathcal{X}$. $\mathbf{x}$ is an $8 \times 8$ pixel cell, and y is the entity corresponding to Y on x. $l_{p}$ is a cross-entropy loss over the cells $\mathbf{x} \in \mathcal{X}$:
\begin{equation}
    l_{p}\left(\mathbf{x} ; y\right)=-\log \left(\frac{\exp \left(\mathbf{x}_{y}\right)}{\sum_{k=1}^{65} \exp \left(\mathbf{x}_{k}\right)}\right).
\end{equation}

The Descriptor Decoder first obtains a descriptor $\mathcal{D}$ with dimension $\frac{H}{8} \times \frac{W}{8} \times D$ through the Conv layer, which reduces the training memory overhead and runtime. After that, the bilinear interpolation and L2 normalization are used to obtain the uniform length descriptor, which has dimension $H\times W\times D$. The descriptor loss is defined as:
\begin{equation}
    \mathcal{L}_d(\mathcal{D},\mathcal{D'},S)=
\frac{1}{(H \times W)^2}\sum\sum l_d(\mathbf{d},\mathbf{d'};s),
\end{equation}
where $\mathcal{D'}$ is the labeled descriptor corresponding to D, $\mathbf{d}$ and $\mathbf{d'}$ are descriptors for cells on image pairs. $S$ is an indicator matrix indicating whether a $\mathbf{d}$ has correspondence with $\mathbf{d'}$. $s$ is a cell-specific indication of the relationship.

In the context of self-supervised training, our approach maintains the same training strategy as SuperPoint by making straightforward geometric transformations to the images. This process yields novel image pairs, accompanied by self-labeled keypoints. Subsequently, these image pairs, characterized by known positional relationships, are fed into the SuperPoint framework to facilitate the extraction of feature points and descriptors. Then the feature points within these image pairs can be systematically matched using the k-Nearest Neighbors (kNN) algorithm.
Finally, the models are jointly trained by iteratively optimizing the two modules at the epoch level with their respective loss functions.

\subsection{Inference}
Homography plays a crucial role in capturing and establishing the relationship between two images of the same scene from different viewpoints.
We have obtained the matching relationship of  keypoints between UAV aerial images and satellite image patches. By fitting a mathematical model, the homography transformation between pairs of images can be estimated. With a set of corresponding points $C_l = (x_1, y_1)$ and $C'_l =(x_2, y_2)$, the homography maps them in the following manner:
\begin{equation}
    \left[\begin{array}{c}
x_{1} \\
y_{1} \\
1
\end{array}\right]= H \left[\begin{array}{c}
x_{2} \\
y_{2} \\
1
\end{array}\right]=\left[\begin{array}{lll}
h_{00} & h_{01} & h_{02} \\
h_{10} & h_{11} & h_{12} \\
h_{20} & h_{21} & h_{22}
\end{array}\right]\left[\begin{array}{c}
x_{2} \\
y_{2} \\
1
\end{array}\right] .
\label{equ:homography}
\end{equation}
The homography matrix $H$ is a $3 \times 3$ matrix but has 8 DoFs because it is usually normalized such that $h_{22}=1$.  Since the equation \ref{equ:homography} holds for all sets of correspondences in the same plane, as long as there are more than 4 sets of correspondences between pairs of images, we can robustly fit $H$ in an optimal way for all correspondences.

As shown in the Inference phase on the right side of Figure \ref{figframework}, we can get the matrix $H$ by matching the keypoints between the UAV aerial image (left) and the best-matched satellite image patch (right). Since the UAV aerial image is orthorectified, the center point $C_c = (x_c, y_c)$ of the image represents the current UAV position, so we obtain the predicted point $C_p = (x_p, y_p)$ in the satellite image:
\begin{equation}
    \left [  x_p \quad y_p \quad 1  \right ]^T = H \left [  x_c \quad y_c \quad 1  \right ]^T .
\end{equation}
Subsequently, the coordinate $p_{i}$ of $C_p = (x_p, y_p)$ can be calculated according to $LL_{coords}$ and $UR_{coords}$, so as to achieve visual localization.

\section{EXPERIMENTS}
After introducing the datasets and implementation details (§\ref{sec:dataset}, §\ref{sec:implementation}), we demonstrate the excellence and generalization of our method in six scenes with varying degrees of texture richness (§ \ref{mainResults}). With extensive ablation studies, we substantiate the capabilities  of the Large-scale Retrieval and Fine-grained Matching modules.  Furthermore, we assess the efficacy of joint optimization of the model during its training process (§ \ref{ablationStudy}).

\subsection{Datasets}
\label{sec:dataset}
Our proposed Large-scale Retrieval module is trained on the Visual Terrain Relative Navigation (VTRN) dataset from the GPR competition\footnote{\url{https://sites.google.com/andrew.cmu.edu/gpr-competition/datasets}}, while the Fine-grained Matching module leverages the RSSDIVCS \cite{li2021learning} dataset for training. 

\begin{figure}[h]
    \centering
    \includegraphics[width=0.9\linewidth]{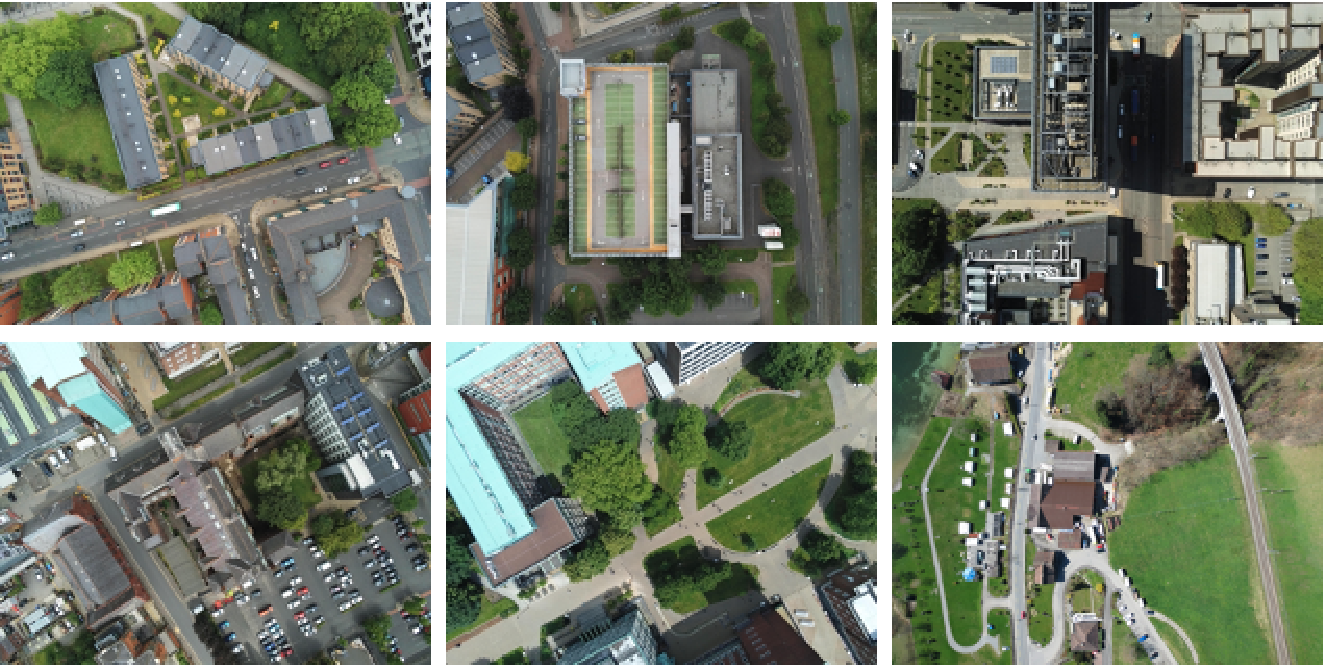}
    \caption{Six UAV aerial scenes. The first row from left to right is Metropolitan University, Energy Center and Business School, the second row is Holy Name Church, Museum and Village.}
    \label{fig:scenes}
\end{figure}

In the UAV visual localization task, the availability of public datasets containing both orthorectified remote sensing images and UAV aerial images accompanied by precise geographic coordinates is notably limited. 
To demonstrate the generalization ability of the proposed model, we employ six distinct scenes of UAV flight with geographic information and display sample images in Figure \ref{fig:scenes}.   The Village scene and corresponding remote sensing image are from \cite{goforth2019gps} to ensure experimental fairness. All other scenes are selected from Manchester Surface Drone Imagery (MSDI)\footnote{\url{https://zenodo.org/record/6977602}} dataset and orthorectified remote sensing images are obtained from Google Earth Pro\footnote{\url{https://www.google.com/earth/versions/#earth-pro}} at a resolution of 1 meter per pixel. A comprehensive detail of datasets is shown in Table \ref{dataset}.

\begin{table}[h]
\renewcommand{\arraystretch}{1.2}
\caption{Dataset details. Underlines indicate the number of query images and wavy lines indicate the number of reference images.}
\label{dataset}
\footnotesize
\centering
\resizebox{\linewidth}{!}{
\begin{tabular}{ccccc}
\toprule[1.5pt]
                           & Dataset       & \multicolumn{2}{c}{Images} & Size        \\ \hline
\multirow{3}{*}{\begin{tabular}[c]{@{}c@{}}Large-scale\\ Retrieval\end{tabular}}        & VTRN Train      & {\uline {10436}}    & \uwave{2853}    & \multirow{3}{*}{500$\times$500} \\
                           & VTRN Val       & {\uline {1684}}          & \uwave{459}           &             \\
                           & VTRN Test      & {\uline {1532}}          & \uwave{1532}         &             \\ \cline{2-5} 
\multirow{2}{*}{\begin{tabular}[c]{@{}c@{}} Fine-grained \\ Matching\end{tabular}} & RSSDIVCS Train & \multicolumn{2}{c}{42000} & \multirow{2}{*}{256$\times$256} \\
                           & RSSDIVCS Test & \multicolumn{2}{c}{14000}           &             \\ \hline
\multirow{6}{*}{Inference} 
& Metropolitan University   & \multicolumn{2}{c}{29}  & 4579$\times$3427 \\
& Energy Center   & \multicolumn{2}{c}{71}  & 4579$\times$3427 \\ 
& Business School   & \multicolumn{2}{c}{37}  & 4579$\times$3427 \\ 
& Holy Name Church  & \multicolumn{2}{c}{58}  & 4579$\times$3427 \\  
& Museum   & \multicolumn{2}{c}{48}  & 4579$\times$3427 \\ 
& Village      & \multicolumn{2}{c}{17}         & 4608$\times$3456 \\

\bottomrule[1.5pt]
\end{tabular}%
}
\end{table}

\subsection{Implementation Details}
\label{sec:implementation}
In order to jointly train our Large-scale Retrieval and Fine-grained Mathcing, we use the layers up to and including conv2\_x of ResNet50 pre-trained on ImageNet \cite{deng2009imagenet} as the shared encoder. All layers after conv2\_x of ResNet50 are used as the backbone of the Large-scale Retrieval. 
All experiments are performed on two NVIDIA GeForce RTX 4090 GPUs. We use the Pytorch framework to implement the proposed network,  both modules of the training network employ Adam optimizer with a learning rate of 0.0001.

\subsection{Main Results}
\label{mainResults}

\begin{figure}[!h]
\centering
	\includegraphics[width=0.8\linewidth]{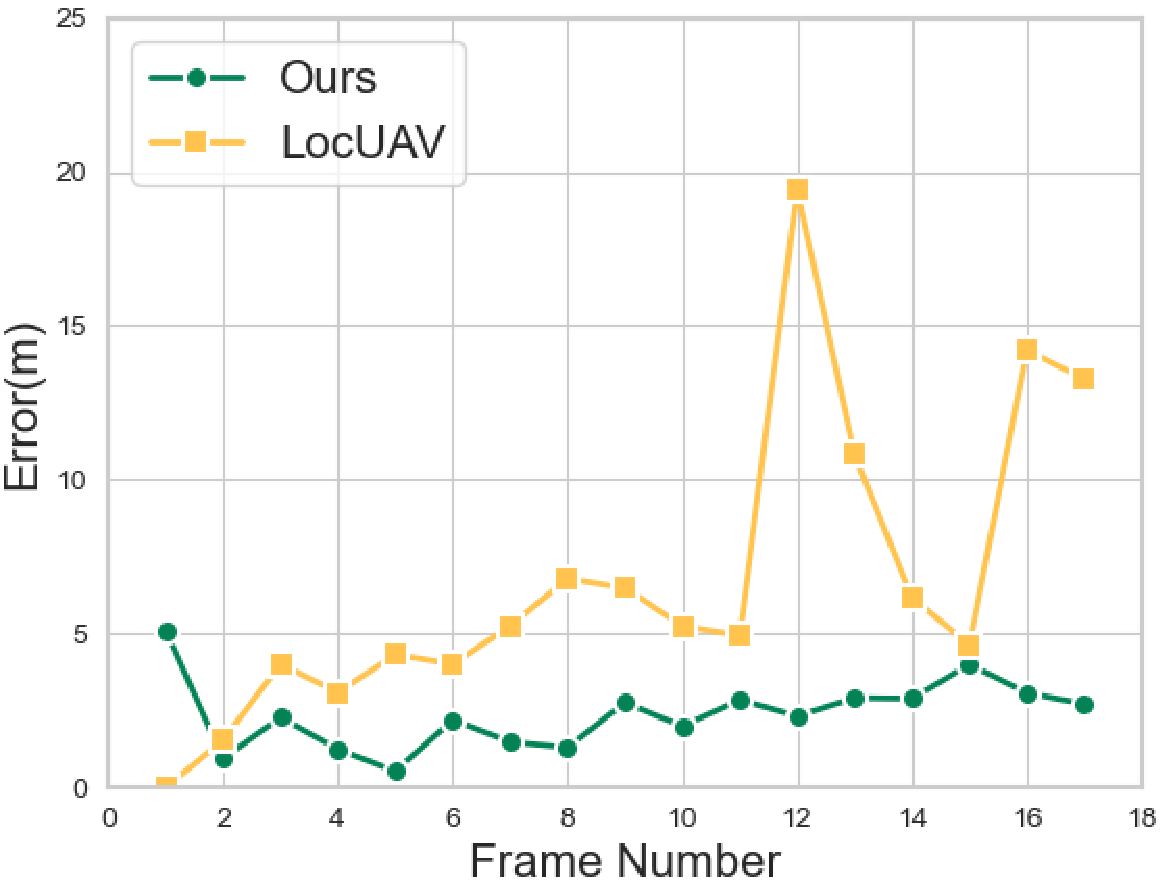}
	\caption{Comparison with baseline on the Village dataset.}
    \label{fig:line_chart}
\end{figure}

As illustrated in Figure \ref{fig:line_chart}, we conducted a comparative analysis of the Village scene in contrast to a state-of-the-art approach LocUAV \cite{goforth2019gps}. Our method achieves an average localization error of $2.39$m, which is considerably better than LocUAV with an error $7.06$m. 

Notably, the experimental configuration of LocUAV \cite{goforth2019gps} involves the given geographic location of the initial frame of the UAV flight, followed by frame-by-frame prediction and localization facilitated through a sliding window mechanism. In contrast, our approach leverages the model's capabilities in image retrieval and feature extraction to achieve autonomous localization, without prior knowledge of the UAVs' initial motion parameters and flight direction. 
The stability and computational efficiency of the homography transformation method grant us the capacity for high-precision keypoint matching, thereby circumventing the need for complex function fitting processes. 

\begin{figure}[!h]
\centering
	\includegraphics[width=0.8\linewidth]{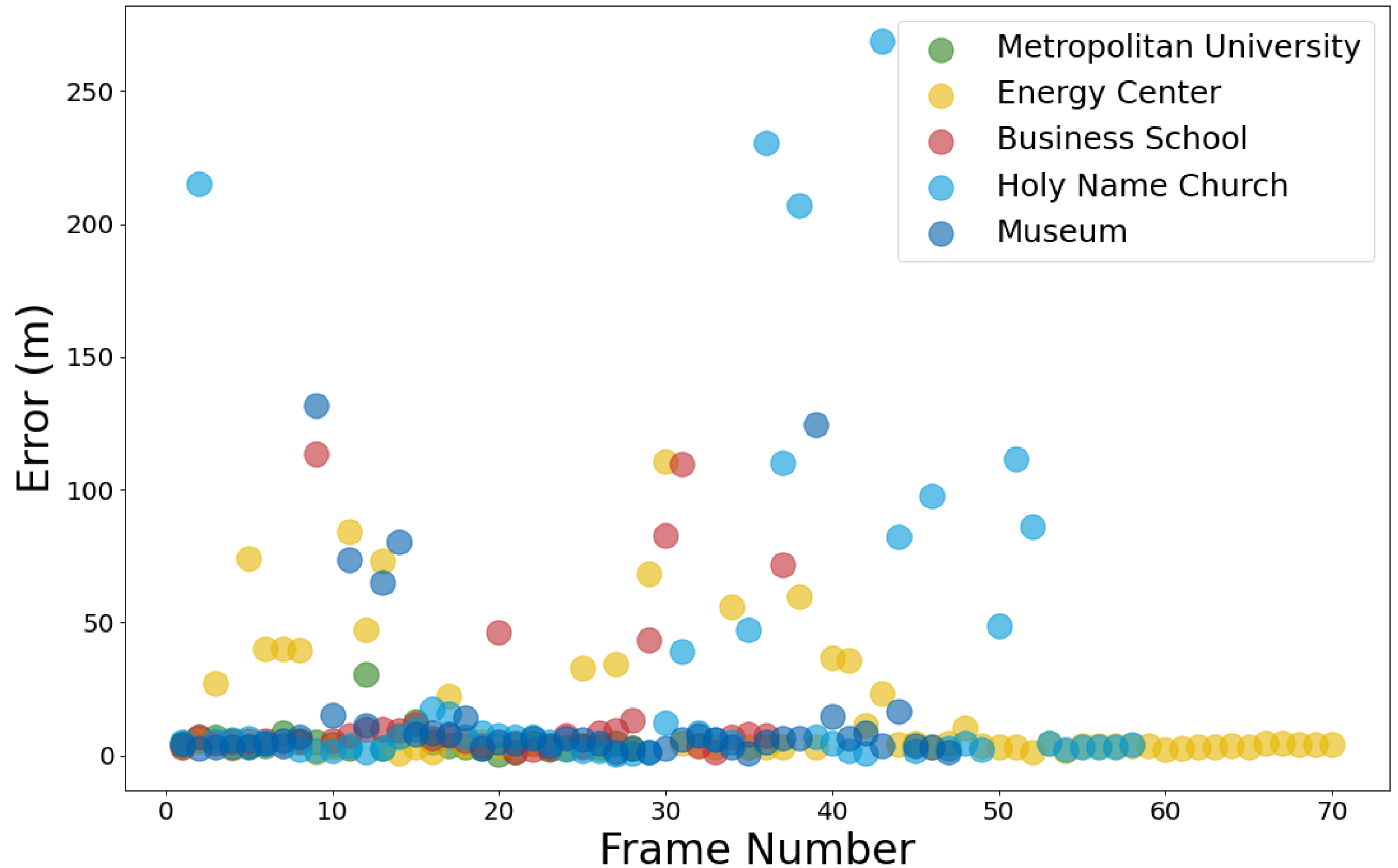}
	\caption{Localization errors for five scenes on the MSDI dataset.}
    \label{fig:scatterplot}
\end{figure}

\begin{table*}[t]\renewcommand{\arraystretch}{1.9}
\centering
\caption{Experimental results and ablation studies on six scenes.  ALE is Average Localizaiton Error, Time refers to the time consumed to localize each frame. R, M and J refer to Large-scale Retrieval module, Fine-grained Matching module and Jointly Optimized Training process respectively. $\mathcal{S}$  denotes that we use the SIFT algorithm instead of our matching module}
\label{ablationResults}
\begin{tabular}{ccccccccccccccc}
\toprule[1.5pt]
\multicolumn{3}{c}{} &
  \multicolumn{2}{c}{Metropolitan University} &
  \multicolumn{2}{c}{Energy Center} &
  \multicolumn{2}{c}{Business School} &
  \multicolumn{2}{c}{Holy Name Church} &
  \multicolumn{2}{c}{Museum} &
  \multicolumn{2}{c}{Village} \\ \cline{4-15} 
R & M                     & \multicolumn{1}{l}{J} & ALE   & Time & ALE   & Time & ALE            & Time & ALE    & Time & ALE   & Time & ALE           & Time \\ \hline

  \checkmark &  $\mathcal{S}$     &        & 73.12 & 1.01 & 99.03 & 0.92 & 70.56          & 0.76 & 45.48  & 0.97 & 69.38 & 0.98 & 60.56         & 1.24 \\
  \checkmark & \checkmark  &                       & 40.93 & 3.72 & 34.15 & 3.66 & 173.51         & 2.19 & 106.78 & 3.62 & 92.15 & 3.67 & 237.94        & 2.29 \\
  & \checkmark &  \multicolumn{1}{l}{\checkmark}                    & 7.03  & 0.55 & 14.25 & 0.62 & \textbf{14.88} & 2.05 & 41.82  & 0.90 & 35.96 & 0.56 & \textbf{1.99} & 3.47 \\
  &   $\mathcal{S}$      &        & 70.89 & 1.48 & 67.77 & 1.62 & 42.68          & 6.16 & 50.01  & 2.51 & 89.51 & 1.44 & 61.01         & 4.41 \\ 
  \checkmark & \checkmark &  \checkmark (Ours)                  &
  \textbf{5.35} &
  \textbf{0.48} &
  \textbf{13.65} &
  \textbf{0.23} &
  17.83 &
  \textbf{0.51} &
  \textbf{30.47} &
  \textbf{0.69} &
  \textbf{15.32} &
  \textbf{0.48} &
  2.39 &
  \textbf{0.59} \\
\bottomrule[1.5pt]
\end{tabular}%

\end{table*}
To the best of our knowledge, our study represents a pioneering effort in addressing the UAV visual localization task on the challenging MSDI dataset. 
Despite the dataset's richness in image texture features, Manchester contains structurally similar buildings, which present a considerable challenge for our model's ability to perform large-scale retrieval. Additionally, the limited altitude of the UAV during image capture results in slight perspective shifts in the photographs of tall buildings, as these images are not orthorectified. 
Moreover, the temporal misalignment between UAV image acquisition and the production of satellite remote sensing images, owing to the rapid urban development in Manchester, further complicates our localization task.
Our experimental results are comprehensively summarized in Table \ref{ablationResults}, with performance evaluated using two metrics: Average Localization Error (ALE) and localization Time per frame.

Our GLVL relies solely on computer vision technology, devoid of any sensor-based or auxiliary error correction mechanisms. Consequently, certain inaccuracies stemming from potential errors in image retrieval and matching processes are inevitable. These inaccuracies can be particularly visible in the form of outliers, which can impact the ALE.
To provide a visual representation of the localization error distribution across the selected MSDI scenes, we present a scatter plot in Figure \ref{fig:scatterplot}. This plot serves as an illustrative representation of the localization error per frame, demonstrating that GLVL does not exert any discernible influence on subsequent localization prediction results, even in instances marked by pronounced localization errors.

\subsection{Ablation Study}
\label{ablationStudy}

\begin{table}[h]
\renewcommand{\arraystretch}{1.3}
\caption{The effect of Jointly Optimized Training on the retrieval module and the experimental metric is recall at K (R@K)}
\label{tab:retrieval}
\begin{tabular}{lllll}
\toprule[1.5pt]
                      & R@1  & R@5  & R@10 & R@20 \\ \cline{2-5} 
Ours                  & \textbf{80.8} & \textbf{88.7} & \textbf{90.9} & \textbf{93.4} \\
Non Jointly Optimized Training & 80.2 & 85.7 & 88.8 & 92.3 \\ 
\bottomrule[1.5pt]
\end{tabular}%
\end{table}

To measure the influence of each model component on UAV visual localization, we designed an ablation study on all inference datasets. The following different settings are displayed in Table \ref{ablationResults} respectively: use our proposed retrieval module, while employing SIFT features in the fine-grained matching process~(row 1); use our proposed retrieval module and the fine-grained matching module, while without joint optimizing~(row 2); without retrieval, directly match image pairs with the joint optimizing ~(row 3); direct matching on all image patches with SIFT features ~(row 4).

\textbf{w/o Large-scale Retrieval module}.
By comparing row 3 and row 5 in Table \ref{ablationResults}, it is interesting to see that the large-scale retrieval module improves both the time consumption and the accuracy.
The global matching policy is not affected by retrieval errors. Consequently, when assessed against the Village and Business School scenes, our methodology exhibits a slightly reduced positioning accuracy in comparison to the only matching Strategy. It results in a mere reduction of $0.4$m in error for the Village dataset and $2.95$m for the Business dataset, an error magnitude less than three pixels.
Conversely, the expansion of the search space increases the matching process and the matching error rate, leading to a significant increase in ALE in other scenes.

\textbf{w/o Fine-grained Matching module}.
By comparing row 1, row 4 and row 5 in Table \ref{ablationResults}, we notice a significant improvement in the localization accuracy of the Fine-grained Matching module.
SIFT's reliance on gradient histograms results in significant computational overhead, necessitating the calculation of gradients for each pixel within the block. 
Consequently, its suitability for low-power devices, such as UAVs, is questionable. The empirical findings from our experimentation reveal a substantial disparity in execution time between this ablation experiment and our proposed approach.
Furthermore, the SIFT algorithm exhibits limitations in handling scenarios involving changes in illumination and image blurring, posing challenges in the alignment of dissimilar image types, notably UAV aerial images and satellite remote sensing images. 

\textbf{w/o Jointly Optimized Training}. 
By comparing row 2 and row 5 in Table \ref{ablationResults}, we found that the results of this ablation experiment almost exhibit the poorest performance with regard to ALE and inference time.
This can be attributed to the increased time overhead incurred by the independent execution of the two-stage model's inference process. 
The adoption of joint training facilitates the shared encoder's exposure to a more extensive training dataset, consequently enhancing the synergy between the two modules and augmenting their respective performances.

In addition, we evaluate the effect of Jointly Optimized Training on the retrieval module. The experimental results are shown in Table \ref{tab:retrieval}, proving the validity of our method.

\section{CONCLUSIONS}

In this work, we provide a novel solution for real-time localization of UAVs. Our approach GLVL combines large-scale retrieval and fine-grained matching into end-to-end joint training, which is capable of focusing on the similarity of local structural features as well as detecting and describing keypoint information.
In six UAV flight scenes encompassing both texture-rich and texture-sparse regions, we demonstrate that GLVL is capable of real-time accurate UAV localization.

\addtolength{\textheight}{-2cm}   




\section*{ACKNOWLEDGMENT}

This work has been partially funded by the Fundamental Research Funds for the Central Universities with number 2023RC61, and the National Natural Science Foundation of China under
Grant 62301063.


\bibliographystyle{IEEEtran}

\bibliography{references}

\end{document}